\documentclass[letterpaper, 10 pt, conference]{ieeeconf}  

\IEEEoverridecommandlockouts                              
\let\labelindent\relax
\usepackage{amsmath}
\usepackage{amsfonts}
\usepackage[graphicx]{realboxes}
\usepackage{graphicx}
\usepackage{graphicx}
\usepackage{textcomp}
\usepackage{booktabs}
\usepackage{fancyhdr}
\usepackage{multirow}
\usepackage{color}
\usepackage{bm}
\usepackage{enumerate} 
\usepackage{enumitem}
\usepackage{mathrsfs}
\usepackage{makeidx}           
\usepackage{multicol}      
\usepackage{psfrag}

\usepackage[ruled,vlined]{algorithm2e}
\usepackage{algorithmic}
\usepackage{colortbl}  
\usepackage[dvipsnames]{xcolor}
\usepackage{array} 

\usepackage{flushend}

\title{\LARGE \bf
I3DOD: Towards Incremental 3D Object Detection via Prompting 
}
\author{Wenqi Liang$^{1,2,3,\dagger}$, Gan Sun$^{1,2,*}$, Chenxi Liu$^{1,2,3,\dagger}$, Jiahua Dong$^{1,2,3}$ and Kangru Wang$^{4}$ \\
{\tt\small \{liangwenqi0123,sungan1412,liuchenxi0101,dongjiahua1995\}@gmail.com~~wangkangru@mail.sim.ac.cn}
\thanks{$^{1}$State Key Laboratory of Robotics, Shenyang Institute of Automation,
Chinese Academy of Sciences, Shenyang, 110016, China;}
\thanks{$^{2}$Institutes for Robotics and Intelligent Manufacturing, Chinese Academy
of Sciences, Shenyang, 110169, China;}
\thanks{$^{3}$University of Chinese Academy of Sciences, Beijing, 100049, China;}
\thanks{$^{4}$Institute of Microsystem and Information Technology, Chinese Academy of Sciences, Shanghai 200050, China.}
\thanks{$^{*}$The corresponding author is Prof. Gan Sun.}
\thanks{$^{\dagger}$These authors contributed equally to this work.}
\thanks{This work is supported by National Nature Science Foundation of China under Grant (62003336, 62273333), CAS-Youth Innovation Promotion Association Scholarship under Grant 2023207, and the State Key Laboratory of Robotics (2022-Z06).} 
}

\begin{document}

\maketitle
\thispagestyle{empty}
\pagestyle{empty}

\begin{abstract}
3D object detection have achieved significant performance in many fields, \emph{e.g.}, robotics system, autonomous driving, and augmented reality. However, most existing methods could cause catastrophic forgetting of old classes when performing on the class-incremental scenarios. Meanwhile, the current class-incremental 3D object detection methods neglect the relationships between the object localization information and category semantic information, and assume all the knowledge of old model is reliable. To address the above challenge, we present a novel \underline{I}ncremental \underline{3D} \underline{O}bject \underline{D}etection framework with the guidance of prompting, \emph{i.e.}, I3DOD. Specifically, we propose a task-shared prompts mechanism to learn the matching relationships between the object localization information and category semantic information. After training on the current task, these prompts will be stored in our prompt pool, and perform the relationship of old classes in the next task. Moreover, we design a reliable distillation strategy to transfer knowledge from two aspects: a reliable dynamic distillation is developed to filter out the negative knowledge and transfer the reliable 3D knowledge to new detection model; the relation feature is proposed to capture the responses relation in feature space and protect plasticity of  the model when learning novel 3D classes. To the end, we conduct comprehensive experiments on two benchmark datasets and our method outperforms the state-of-the-art object detection methods by $0.6\% \sim 2.7\%$ in terms of mAP@0.25.

\end{abstract}

\definecolor{deepred}{rgb}{0.698,0.133,0.133}
\section{INTRODUCTION}
 3D object detection has recently received widespread attention in computer vision and robot perception fields, which is developed to localize and identify 3D objects in a scene. It performs a vital role in autonomous driving \cite{b4}, robotics system \cite{b11} and augmented reality \cite{b13}. Especially in robotics systems, 3D object detection can help the robot to perform 3D scene understanding and object location. For instance, robots can effectively carry out the task of object grasp, obstacle avoidance and context awareness with the assistance of 3D object detection,.
 
\begin{figure}[t]
	\centering
	\includegraphics[width=245pt,height=110pt]
	{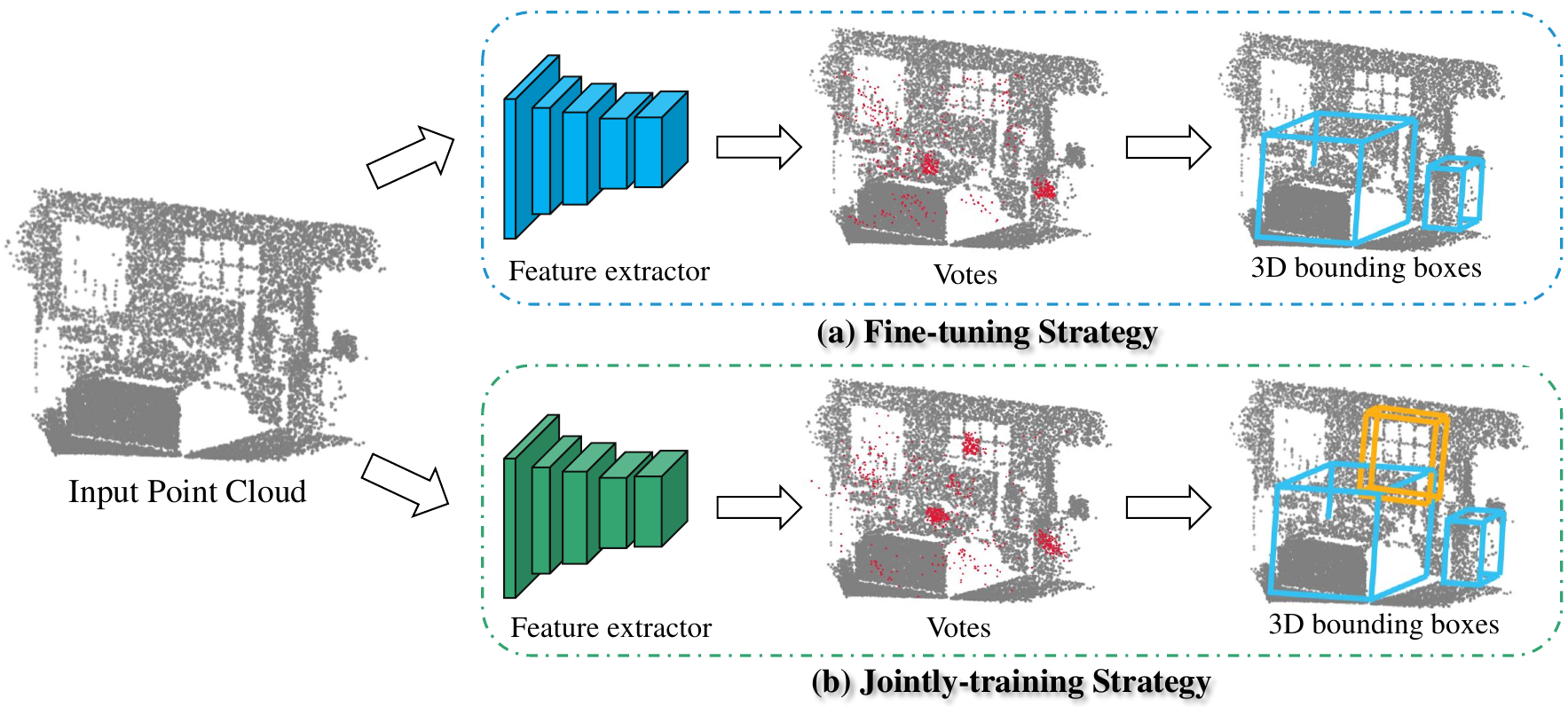}
	\vspace{-20pt}
	\caption{Qualitative detection results obtained from: (a) fine-tuning on the new 3D classes and (b) jointly-training strategy on all the 3D classes, where the \textcolor{red}{\textbf{red}} points denotes the votes generated by VoteNet \cite{b6}, and \textcolor{orange}{\textbf{yellow}} and \textcolor{blue}{\textbf{blue}} in the output denotes the bounding boxes of old 3D classes and the new 3D classes, respectively.} 
	\label{fig: motivation}
	\vspace{-15pt}
\end{figure}

 However, the current deep learning-based methods (\emph{e.g.}, \cite{b6} \cite{b8}) in 3D object detection are trained in the invariable data, while the data of the real world is constantly changing through time. For example, a domestic robot can detect the common domestic garbage in the indoor scene, and then grasp to clean it up. When the owner brings garbage with a brand-new package that the robot has never seen before,  one straightforward way for the robot to detect the new garbage is by fine-tuning our model on the point cloud data of new classes. However, this manner will lead to the notorious catastrophic forgetting of old 3D objects. The other way is adopting the existing class-incremental 2D object detection methods to the 3d scenes, \emph{e.g.}, Zhao \emph{et al.} \cite{b2} propose the co-teaching method SDCoT to distill 3D knowledge from old model. However, the SDCoT model could neglect the object localization information, when assuming that the way of transferring the anti-forgetting knowledge of 2D objects is identical with 3D objects. As shown in Fig.~\ref{fig: motivation}, given an input point cloud, object detection model can locate the object to predict 3D bounding boxes and category information after capturing the center of object from the high-level feature (\emph{i.e}. red points). Failure to match this relationship between the object localization information and category semantic information exacerbates catastrophic forgetting in old classes. Moreover, when performing knowledge distillation, current methods are convinced that all the responses produced by old model are positive for new model, whereas Feng \emph{et al.} \cite{b15} have proved that the negative responses exist and could affect the performance of new model.
 
\begin{figure*}[t]
	\centering
	\includegraphics[width=490pt, height=170pt]
	{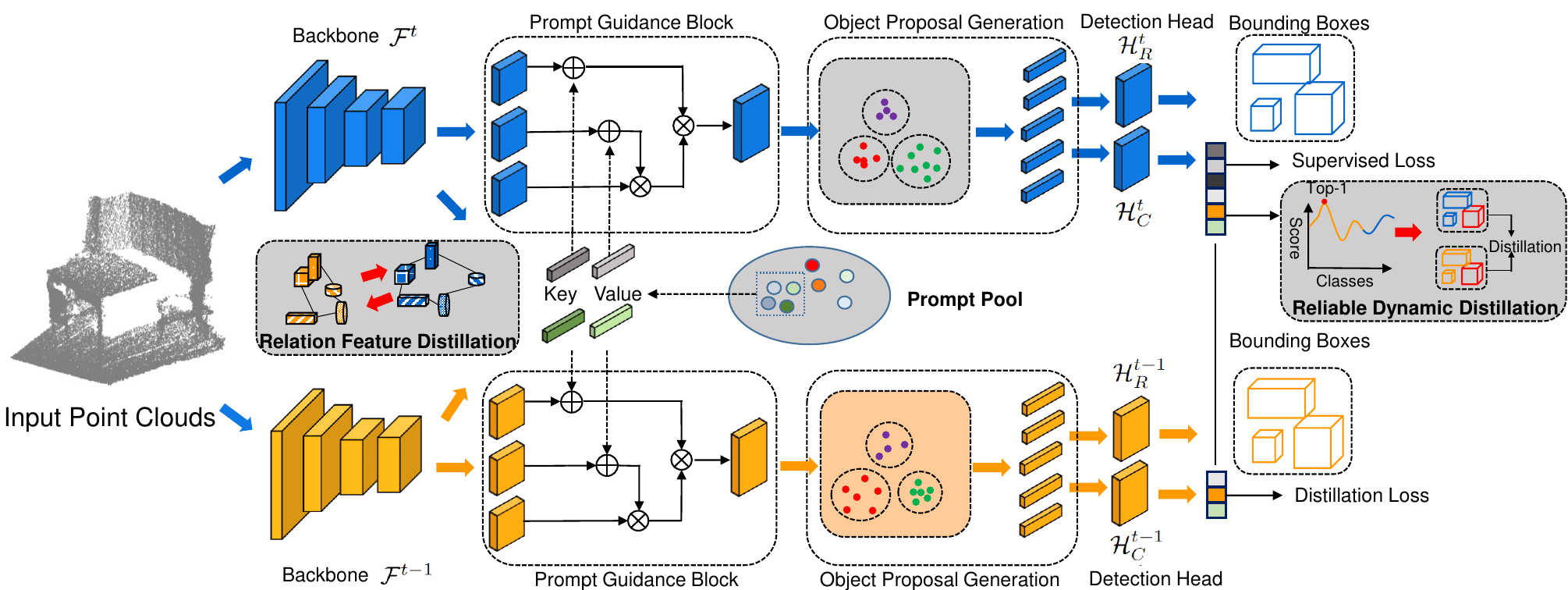}
	\vspace{-10pt}
	\caption{Overview framework of our method I3DOD, whose network structure  is based on VoteNet \cite{b6}. It mainly consists of a Prompt Guidance Block to learn the category space information, a Reliable Dynamic Distillation module to screen out the reliable 3D knowledge from the regression head, and a Relation Feature Distillation module to distill the spatial positional relation in feature space.} 
	\label{fig: overview_of_our_model}
	\vspace{-15pt}
\end{figure*}

To address the above problems, we propose a novel \underline{I}ncremental \underline{3D} \underline{O}bject \underline{D}etection framework with the guidance of prompting, \emph{i.e.}, I3DOD. Specifically, we present two aspects of comprehension about the challenge of catastrophic forgetting. To reinforce the above-mentioned relationship of each class, we explore a prompt-based method. Concretely, we initialize some prompts to capture the the object localization information from the high-level semantic information. Furthermore, these prompts are stored in our prompt pool and updated with learning a series of tasks. When trained in the next task, our model can adopt the task-shared prompts to perform the above-mentioned relationship of old classes. Moreover, we apply a reliable distillation strategy to distill the reliable 3D knowledge from old model to new model. The reliable dynamic distillation is designed to filter out the negative knowledge produced by the regression head of old model. In addition, we develop the relation feature distillation to protect plasticity of new model, and distill the distribution of responses in feature space.

In the end, the main contributions of this paper are presented as follows: 

\begin{itemize}
\item We propose a novel class-incremental 3D object detection method with the guidance of prompting, which could learn the new 3D object in a plasticity manner while mitigating the damage of catastrophic forgetting. 

\item A prompt guidance block is designed to store prompts including the comprehension of old category information, which could explore the relationship between the object localization information and high-level semantic information in class-incremental 3D object detection. 

\item  A reliable distillation strategy is designed to address heterogeneous forgetting, which can purify the anti-forgetting 3D knowledge from old model and protect the plasticity of new model. Our method achieves significant performance on SUN RGB-D and ScanNet in comparison to the state-of-the-art.

\end{itemize}

\section{Related Work}

\subsection{Class-Incremental learning}

Since the issue of class-incremental learning was raised more than 20 years ago \cite{b3}, class-incremental learning \cite{b17,b19, ma2023smca} has received much attention in different fields.  Following \cite{b5}, we classify class-incremental learning into four types: regularisation-based \cite{b9,b7}, replay-based \cite{b10,b12}, pseudo-sample generation \cite{b16,b14}, and architecture-based \cite{b18,ICCV2023_HFC,b21}. To distill knowledge of old tasks, knowledge distillation is introduced into incremental learning by LWF \cite{b7}. Recently, DDE \cite{b25} demonstrates the importance of causal effect to resist catastrophic forgetting and introduces it to knowledge distillation.
In 2D object detection, the early method \cite{b23} firstly applies knowledge distillation on the faster-RCNN architecture. Faster ILOD \cite{b29} further adopts an adaptive multi-network distillation loss that enhances the accuracy and efficiency of the RPN-based detection method. MVCD \cite{b24} extends the application of knowledge distillation to further retain knowledge of old tasks by retaining correlations at the channel, point and instance levels. Class-incremental learning for 3D object detection is still under development. In \cite{b2} , SDCoT introduces knowledge distillation into 3D class-incremental object detection to guarantee the performance of model on old classes. However, the above-mentioned methods neglect the object localization information, and assume that all responses produced by old model are reliable for new model.
\subsection{Prompt-based learning}

Prompt-based learning is first proposed in natural language processing to control the model prediction output and save the cost of pre-training. L2P \cite{b26} introduces prompt-based learning into class-incremental learning to guide model prediction with less computational cost, as prompt-based learning is able to capture task-specific knowledge with smaller additional parameters. In \cite{b30}, DualPrompt further improves the performance by adding complementary learning systems to prompt-based learning. However, there still exists difficulties when extends the prompt-based learning into 3D object detection.

\section{The Proposed Method}

\subsection{Problem Definition and Method Overview} According to class-incremental 3D object detection, the model observes a series of class-incremental learning tasks $\{\mathcal{T}^t\}_{t=1}^T$ and a data stream $\{\mathcal{D}^t\}_{t=1}^T$, where $T$ represents the total number of tasks. Following SDCoT, the training data  $\mathcal{D}^t$ and the class set $\mathcal{C}^t$ are only available for the $t$-th task, \emph{i.e.} $\mathcal{D}^t\cap(\cup_{i=1}^{t-1}\mathcal{D}^i)= \emptyset$ and $\mathcal{C}^t\cap(\cup_{i=1}^{t-1}\mathcal{C}^i) = \emptyset$. Given a stream $\{\mathcal{D}^t, \mathcal{C}^t\}_{t=1}^T$ in the $t$-th task, we aim to detect 3D objects from all current classes $\mathcal{C}^t$ and old classes $\cup_{i=1}^{t-1}\mathcal{C}^i$.

We present the overview of our I3DOD in Fig.~\ref{fig: overview_of_our_model}. A core purpose of our method is to tackle catastrophic forgetting of previous knowledge, when the training data $\cup_{i=1}^{t-1}\mathcal{D}^I$ is not available for the current task. Specifically, we define the backbone and model in the current task as $\mathcal{F}^t$ and $\Phi^t$. The old backbone $\mathcal{F}^{t-1}$ and $\Phi^{t-1}$ are learned from the last task. The head of each model is composed of a regression head $\mathcal{H}_{R}$ and a classification head $\mathcal{H}_{C}$. To tackle the forgetting of old classes without affecting the learning of new classes, we propose our solution from two aspects. On the one hand, we propose a prompt guidance block to record the relation between the category semantic information of each class and its the object localization information, which can guide the matching of the two pieces of information to solve catastrophic forgetting. On the other hand, we design a distillation scheme (\emph{i.e.} $\mathcal{L}_{RFD}$, $\mathcal{L}_{RBD}$ and $\mathcal{L}_{Dis}$) in Fig.~\ref{fig: overview_of_our_model} to effectively extract knowledge of old model.

\subsection{Prompt Guidance Learning}   
   In class-incremental 3D object detection, SDCoT \cite{b2} introduces a co-teaching distillation to transfer knowledge between two teacher models and a student model. However, this method is very common in 2D vision tasks and neglects the matching relationship between the object localization information and high-level semantic information. As shown in Fig.1, we find that a matching relationship between the two pieces of information of old classes is crucial for the new model to capture the center of object and locate it for object detection. Inspired by L2P \cite{b26} and  DualPropmt \cite{b30}, we propose a prompt guidance block to address the above issues from the comprehension of two pieces of information.

   As shown in Fig.~\ref{fig: overview_of_our_model},  given an input point cloud $\mathbf{x}_i^t$, a feature vector $\mathbf{f}_i^t$ and a subset of the input points $\mathbf{z}_i^t$ are obtained from the output of $\mathcal{F}^t$:
\begin{align}
	\label{eq: feature output}
    \{\mathbf{z}_i^t,\mathbf{f}_i^t\} = \mathcal{F}^t(\mathbf{x}_i^t),
\end{align}   
where $\mathbf{x}_i^t$ and $\mathbf{z}_i^t \in\mathbb{R}^{3}$,  $\mathbf{f}_i^t\in\mathbb{R}^{M\times D}$.

To build the relationship between the high-level feature $\mathbf{f}_i^t$ and the center of 3D object, we here introduce our prompt pool $\{\mathbf{p}_n\}_{n=1}^N$ , where $N$ represents the total prompt quantity and $\mathbf{p}_i^t \in\mathbb{R}^D$. Inspired by a great success of  transformer, we introduce a multi-head self-attention module as our prompting function to combine the selected task-share prompts $\{\mathbf{p}_s\}_{s=1}^S$ and the high-level feature $\mathbf{f}_i^t$. 
After obtaining $\{\mathbf{p}_s\}_{s=1}^S$ and $\mathbf{f}_i^t$, we feed them into our multi-head prompting self-attention module. Specifically, we obtain $\mathbf{h}_Q$, $\mathbf{h}_K$, $\mathbf{h}_V$ by repeating $\mathbf{f}_i^t$, where $\mathbf{h}_Q$=$\mathbf{h}_K$=$\mathbf{h}_V$. In addition, we divide each  $\mathbf{p}$  to gain $\mathbf{p}_K$ and $\mathbf{p}_V$, where $\mathbf{p}_K$, $\mathbf{p}_V\in\mathbb{R}^{S/2\times D}$. We concatenate $\mathbf{p}_K$, $\mathbf{p}_V$ with  $\mathbf{h}_K$, $\mathbf{h}_V$ and execute parallel self-attention map H times. In the end, we obtain the $h$-th self-attention map $\mathbf{A}_h$ as: 
\begin{align}
	\label{eq: self-attention}
\mathbf{A}^h =  \sigma(\frac{\mathbf{h}_Q \mathbf{W}_Q^h ([\mathbf{p}_K;\mathbf{h}_K] \mathbf{W}_K^h)^{\top}}{\sqrt{d}})([\mathbf{p}_V;\mathbf{h}_V]  \mathbf{W}_V^h),
\end{align}   
where $\mathbf{W}_k^h$,$\mathbf{W}_Q^h$,$\mathbf{W}_V^h$ are the weight matrices of $h$-th self-attention module, and $\sigma$ denotes the softmax function.

\begin{table*}[t]
\centering
\setlength{\tabcolsep}{0.75mm}
\renewcommand\arraystretch{1.5}
\normalsize
\caption{Comparison results on SUN RGB-D dataset in terms of mAP@0.25 and Recall.}
\vspace{-5pt}
\scalebox{0.715}{
\begin{tabular}{c|ccc|c|ccc|c|ccc|c|ccc|c|ccc|c|ccc|c} \hline
	\toprule
	\multirow{3}{*}{Methods} & \multicolumn{12}{c|}{mAP@0.25(\%)}  & \multicolumn{12}{c}{Recall(\%)} \\\cline{2-25}
 & \multicolumn{4}{c|}{5 + 5} & \multicolumn{4}{c|}{7 + 3} & \multicolumn{4}{c|}{9 + 1}   & \multicolumn{4}{c|}{5 + 5} & \multicolumn{4}{c|}{7 + 3} & \multicolumn{4}{c}{9 + 1} \\ 
	& B & N   & Avg. & Imp. & B & N   & Avg. & Imp. & B & N   & Avg. & Imp. & B & N   & Avg. & Imp. & B & N   & Avg. & Imp. & B & N   & Avg. & Imp.\\ 
	\midrule
	Freeze and add &57.8  &28.0  & 42.9 & $\Uparrow$15.1 &54.9  &35.8  & 45.4 & $\Uparrow$10.2 &56.4 &\textcolor{deepred}{\textbf{46.1}}   &\textcolor{deepred}{\textbf{51.3}} & $\Downarrow$2.3  &80.3 &60.2  &70.3   & $\Uparrow$12.1 &80.7 &71.4  &76.1   & $\Uparrow$6.4 &80.3  &\textcolor{deepred}{\textbf{80.0}}   &80.2 & $\Uparrow$0.7   \\
 	Fine-tuning  & 57.8 & 30.5 &44.2  & $\Uparrow$13.8 & 54.9& 21.2 & 38.1  & $\Uparrow$17.5 & 56.4 & 5.6  &31.0 & $\Uparrow$18.0  &80.3  & 65.9 & 73.1 & $\Uparrow$9.3 &80.7 &59.1  & 69.9  & $\Uparrow$12.6 &80.3  &30.7   & 55.5& $\Uparrow$22.0  \\
  	SDCoT \cite{b2} &57.8  & 56.3 & 57.1 & $\Uparrow$0.9 &54.9 &53.8  & 54.4  & $\Uparrow$1.2 &56.4  & 38.6  &47.5 & $\Uparrow$1.5 &80.3  &81.3  &79.4  & $\Uparrow$3.0 &80.7 & 79.4 & 80.1  & $\Uparrow$2.4 &80.3  & 72.7&76.5 & $\Uparrow$4.4   \\
    Ours w/o PGB \& RDD \& RFD &57.8  &54.2  &56.0  & $\Uparrow$2.0 &54.9 &53.3  & 54.1  & $\Uparrow$1.5 &56.4  &36.1   &46.3 & $\Uparrow$2.7 & 80.3 &82.5  &81.4  & $\Uparrow$1.0 &80.7 &81.0  &80.9   & $\Uparrow$1.6 &80.3  &72.1   &76.2 & $\Uparrow$4.7  \\
    Ours w/o PGB \& RDD &57.8  &55.5  &56.7  & $\Uparrow$1.3 &54.9 & 54.3 &54.6   & $\Uparrow$1.0 &56.4  &38.9   &47.7 & $\Uparrow$1.3  &80.3  &81.0  &80.7  & $\Uparrow$1.7 &80.7 & 80.2 &80.5   & $\Uparrow$2.0 & 80.3 &  72.9 &76.6 & $\Uparrow$4.3 \\
    Ours w/o PGB  &57.8  &56.4  &57.1  & $\Uparrow$0.9 &54.9 & 54.6 & 54.8  & $\Uparrow$0.8 &56.4  & 40.0  &48.2 & $\Uparrow$0.8  &80.3  &81.6  &81.0  & $\Uparrow$1.4 &80.7 &81.9 &81.3   & $\Uparrow$1.2 &80.3  &76.8   &78.6 & $\Uparrow$2.3 \\
    Ours  &\textcolor{deepred}{\textbf{58.7}}  &\textcolor{deepred}{\textbf{57.2}}  &\textcolor{deepred}{\textbf{58.0}}   &$\mathrm{-}$ &\textcolor{deepred}{\textbf{56.3}} &\textcolor{deepred}{\textbf{54.8}}  &\textcolor{deepred}{\textbf{55.6}}  &$\mathrm{-}$  &\textcolor{deepred}{\textbf{56.7}} & 41.3  &49.0 &$\mathrm{-}$  &\textcolor{deepred}{\textbf{80.4}} &\textcolor{deepred}{\textbf{84.6}}  &\textcolor{deepred}{\textbf{82.4}}  &$\mathrm{-}$  &\textcolor{deepred}{\textbf{82.5}} &\textcolor{deepred}{\textbf{82.6}}  &\textcolor{deepred}{\textbf{82.5}}  &$\mathrm{-}$ &\textcolor{deepred}{\textbf{84.9}}  &76.9   &\textcolor{deepred}{\textbf{80.9}} &$\mathrm{-}$  \\
	
	\bottomrule
\end{tabular}} \hrule
\label{tab: sunrgb}  
\vspace{-10pt}
\end{table*}

\begin{table*}[t]
\centering
\setlength{\tabcolsep}{1.05mm}
\renewcommand\arraystretch{1.5}
\normalsize
\caption{Comparison results on ScanNet dataset in terms of mAP@0.25 and Recall.}
\vspace{-5pt}
\scalebox{0.74}{
\begin{tabular}{c|ccc|c|ccc|c|ccc|c|ccc|c|ccc|c|ccc|c} \hline
	\toprule
	\multirow{3}{*}{Methods} & \multicolumn{12}{c|}{mAP@0.25(\%)}  & \multicolumn{12}{c}{Recall(\%)} \\\cline{2-25}
 & \multicolumn{4}{c|}{9 + 9} & \multicolumn{4}{c|}{14 + 4} & \multicolumn{4}{c|}{17 + 1}   & \multicolumn{4}{c|}{9 + 9} & \multicolumn{4}{c|}{14 + 4} & \multicolumn{4}{c}{17 + 1} \\ 
	& B & N   & Avg. & Imp. & B & N   & Avg. & Imp. & B & N   & Avg. & Imp. & B & N   & Avg. & Imp. & B & N   & Avg. & Imp. & B & N   & Avg. & Imp.\\ 
	\midrule
	Freeze and add &60.8  &30.8  &45.8 & $\Uparrow$14.1 &53.5  &42.3  & 47.9 &$\Uparrow$7.5   &57.4 & 53.8 & 55.6 & $\Uparrow$0.2  &80.3 &55.0  &67.7   & $\Uparrow$11.1 &76.8 &69.2  &73.0   & $\Uparrow$2.1 & 79.6 &77.1   &78.4  &$\Uparrow$0.2   \\
 	Fine-tuning  &60.8  &27.0  &43.7  & $\Uparrow$16.2 &53.5 &15.7  & 34.6  & $\Uparrow$20.8 &57.4  &1.0   &29.2 & $\Uparrow$26.6  & 80.3 &50.5  &65.4  & $\Uparrow$13.4 &76.8 &36.7  & 56.8  & $\Uparrow$18.3 &79.6  &8.9   &44.3 & $\Uparrow$23.3  \\
  	SDCoT \cite{b2} &60.8  &54.3  &57.6  & $\Uparrow$2.3 &53.5 &55.0  &54.3   & $\Uparrow$1.1 &57.4  &53.0   &55.2 & $\Uparrow$0.6 &80.3 &75.2  &77.8  & $\Uparrow$1.0 &76.8 &74.7  &75.8   & $\Uparrow$1.3& 79.6 &74.6   &77.1 & $\Uparrow$1.5   \\
    Ours  &\textcolor{deepred}{\textbf{61.4}}  &\textcolor{deepred}{\textbf{58.4}}  &\textcolor{deepred}{\textbf{59.9}} &$\mathrm{-}$  &\textcolor{deepred}{\textbf{53.8}} &\textcolor{deepred}{\textbf{56.9}}  &\textcolor{deepred}{\textbf{55.4}} &$\mathrm{-}$  &\textcolor{deepred}{\textbf{57.5}}  &\textcolor{deepred}{\textbf{54.2}}   &\textcolor{deepred}{\textbf{55.8}} &$\mathrm{-}$ &\textcolor{deepred}{\textbf{80.6}}  &\textcolor{deepred}{\textbf{76.9}}  &\textcolor{deepred}{\textbf{78.8}}  &$\mathrm{-}$  &\textcolor{deepred}{\textbf{76.9}}  &\textcolor{deepred}{\textbf{77.2}}   &\textcolor{deepred}{\textbf{77.1}}   &$\mathrm{-}$ &\textcolor{deepred}{\textbf{79.8}}   &\textcolor{deepred}{\textbf{77.3}}   &\textcolor{deepred}{\textbf{78.6}} &\textbf{$\mathrm{-}$}  \\
	
	\bottomrule
\end{tabular}}\hrule
\label{tab: scannet}
\vspace{-15pt}
\end{table*}

\begin{table}[t]
\centering
\setlength{\tabcolsep}{0.6mm}
\renewcommand\arraystretch{1.5}
\normalsize
\caption{Performance of each class on SUN RGB-D dataset under the setting of 5+5.}
\vspace{-5pt}
\hrule
\scalebox{0.72}{
\begin{tabular}{c|ccccc>{\columncolor{Cyan!8}}c>{\columncolor{Cyan!8}}c>{\columncolor{Cyan!8}}c>{\columncolor{Cyan!8}}c>{\columncolor{Cyan!8}}c|c|c} 
	\toprule
	\multirow{2}{*}{Methods/Classes} & \multicolumn{12}{c}{mAP@0.25(\%)}  \\\cline{2-13} 
	& 1 & 2  & 3 & 4 & 5 & 6 & 7 & 8 & 9 &10 & Avg. & Imp. \\ 
	\midrule
	Freeze and add &66.7  &79.1  &26.0 & 69.6 & 20.5 &0.7  & 0.1 &6.0   &9.6 & 1.7 &28.0  &$\Uparrow$29.2 \\
 	Fine-tuning  &0.2  &2.2  &1.0  &2.4  &16.0 &28.5  & 54.5  &62.5 &52.3 
  & 85.8 & 30.5 & $\Uparrow$26.7\\
  	SDCoT \cite{b2} &68.1  &\textcolor{deepred}{\textbf{83.4}}   &26.5  & 71.3 &20.1 &32.6  &58.4   &63.1  &52.0  &\textcolor{deepred}{\textbf{87.1}}    & 56.3& $\Uparrow$0.9 \\
    Ours  &\textcolor{deepred}{\textbf{69.5}}  &82.7  &\textcolor{deepred}{\textbf{29.5}} 
    & \textcolor{deepred}{\textbf{71.8}} &\textcolor{deepred}{\textbf{21.3}} &\textcolor{deepred}{\textbf{34.9}}  &\textcolor{deepred}{\textbf{58.6}} &\textcolor{deepred}{\textbf{64.6}} &\textcolor{deepred}{\textbf{52.6}}  &86.1 &\textcolor{deepred}{\textbf{57.2}} &$\mathrm{-}$    \\
	
	\bottomrule
\end{tabular}} \\
\hrule
\label{tab: class-wise}
\vspace{-15pt}
\end{table}

We concatenate all $\mathbf{A}_h$ and apply the MLP layer and the one-dimensional convolution layer to extract the center information of 3D objects. As mentioned above, we use prompts to learn the relationship between the high-level feature and the object localization information in class-incremental learning, and store them for the next task. The outputs of our prompt guidance block are given as follows:
\begin{align}
	\label{eq: eq3}
    \dot{\mathbf{z}}_i^t=\mathbf{z}_i^t+\mathbf{MLP}(\mathbf{Concat}(\{\mathbf{A}_h\}_{h=1}^H)),
\end{align}  
\begin{align}
    \ddot{\mathbf{z}}_i^t = \mathbf{z}_i^t +\mathbf{Conv}(\dot{\mathbf{z}}_i^t).
	\label{eq: eq4}
\end{align}

Overall, we have established the structure of prompt guidance block. By following VoteNet \cite{b6}, the outputs of prompt guidance block are expected to locate the center of objects. Therefore, for training our prompt guidance block, we compute the center of object as the supervisory label. The loss function is shown as below:
\begin{align}
       \label{eq: eq5}
       \mathcal{L}_{PGB} = \frac{1}{B}\sum_{i=1}^B \Vert 
 \ddot{\mathbf{z}}_i^t - \mathbf{c}_i^t \Vert_1, 
\end{align}
where $\mathbf{c}_i^t$ represents the ground truth of object center.

In addition, the prompts have learned the matching relationship of categories in the current task. We build a prompt pool to store these prompts and perform the relationship to tackle catastrophic forgetting of old classes in the next task. 

\renewcommand{\algorithmicrequire}{\textbf{Input:}}
\renewcommand{\algorithmicensure}{\textbf{Output:}}

\begin{algorithm}[t]			
    	\caption{Optimization pipeline of Our I3DOD.} 
	\label{alg: optimization}
	\begin{algorithmic}[1]
		\REQUIRE The data stream $\{\mathcal{D}^t\}_{t=1}^T$.  The hyper-parameters \{$\alpha, \beta, \gamma, \xi, \zeta$\}. Randomly initialize $\Phi^1$, $\{\mathbf{p}_s\}_{s=1}^S$.
		\FOR {$t=1, 2, \cdots, T$}
        \STATE  Initialize the classifier $\mathcal{H}_C^t$;
        \WHILE{not converged}
          \IF {$t=1$}
          \STATE    Optimize $\Phi^t$, $\{\mathbf{p}_s\}_{s=1}^S$ by $\mathcal{L}_S$;
         \ELSE
          \STATE  Optimize $\Phi^t$, $\{\mathbf{p}_s\}_{s=1}^S$ by Eq. (9);
              		\ENDIF
            		\ENDWHILE
		  \STATE  Store $\{\mathbf{p}_s\}_{s=1}^S$ in prompt pool and $\Phi^t$ as $\Phi^{t-1}$.
		\ENDFOR \\
  
	\end{algorithmic} 

\end{algorithm}

\subsection{Reliable Dynamic Distillation}
   In class-incremental 2D or 3D object detection, performing bounding boxes distillation between old and new object detection models is difficult. Given a sample that does not contain old classes, the regression head of old model also generates bounding boxes at low confidence. There is too much noise in the responses of the regression head when applying old model to predict on training data of the current task. Inspired by \cite{b15}, we can divide all the responses into reliable responses and unreliable responses. Transferring unreliable responses from old model to new model will affect learning new classes, and accelerate catastrophic forgetting of old classes. Therefore, the current method SDCoT \cite{b2} cannot adopt the knowledge distillation strategy on bounding boxes.

    To tackle the problems above and effectively transfer the 3D knowledge of old model, we propose a reliable dynamic distillation module to dynamically screen out the reliable 3D knowledge from the regression head.
    As shown in Fig.~\ref{fig: overview_of_our_model}, through the forward calculation of $\Phi^t$, we can obtain classification scores $\mathcal\{{S}_{i}^t\}_{i=1}^N$ and bounding boxes $\mathcal\{{B}_{i}^t\}_{i=1}^N$, where $N$ means the total quantity of proposals generated by object proposal generation module. Specifically, we initialize classification scores of $\Phi^t$ to our confidence scores for each proposal. Specifically, we select the subset of $\mathcal\{{S}_{i}^t\}_{i=1}^N$ whose top-1 classification score is distributed in old classes. We then compute the average $\mu$ and the standard deviation $\sigma$ of all selected classification scores. The final threshold value $\tau$ is calculated below: 
    \begin{align}
        \label{eq: eq6}
          \tau= \mu + \zeta \cdot\sqrt{\frac{\sum_{i=1}^B(S_{i}^t-\mu)\cdot\mathbb{I}_{\mathrm{argmax} S_{i}^t\in \cup_{i=1}^{t-1}\mathcal{C}^i}}{\sum_{i=1}^B\mathbb{I}_{\mathrm{argmax}S_{i}^t\in \cup_{i=1}^{t-1}\mathcal{C}^i} }},
    \end{align}
where $\mu=\frac{1}{O}\sum_{i=1}^O S_{i}^t\cdot\mathbb{I}_{\mathrm{argmax}S_{i}^t\in \cup_{i=1}^{t-1}\mathcal{C}^i}$, and  $\mathbb{I}$ means the indicator function that $\mathbb{I}=1$ when the condition is true; or $\mathbb{I}=0$. The hyperparameter $\zeta$ is designed to control $\tau$. 

 We obtain bounding boxes $\mathcal\{{B}_{i}^t\}_{i=1}^O$, $\mathcal\{{B}_{i}^{t-1}\}_{i=1}^O$ of proposals  whose confidence score exceeds the threshold $\tau$. These reliable responses can transfer positive 3D knowledge from the teacher model to the student model, which is crucial to prevent catastrophic forgetting. Thus, the proposed reliable dynamic distillation for bounding boxes can be defined as:
\begin{align}
    \label{eq: eq7}
    \mathcal{L}_{RDD} = \frac{1}{O}\sum_{i=1}^O \Vert {B}_{i}^t - {B}_{i}^{t-1}  \Vert_2.
\end{align}
\subsection{Relation Feature Distillation}

Most current CIL distillation methods based on feature \cite{b19,b20} neglect the spatial positional relationship in feature space, while further damaging the model plasticity when learning novel 3D classes. They assume that the direct distillation can exert the same effect in the feature space as it does in the output space. However, the distribution of responses in feature space is complex. Directly distilling responses cannot well transfer knowledge form the distribution. 

  \begin{figure*}[t]
	\centering
	\includegraphics[width=480pt, height=110pt]
	{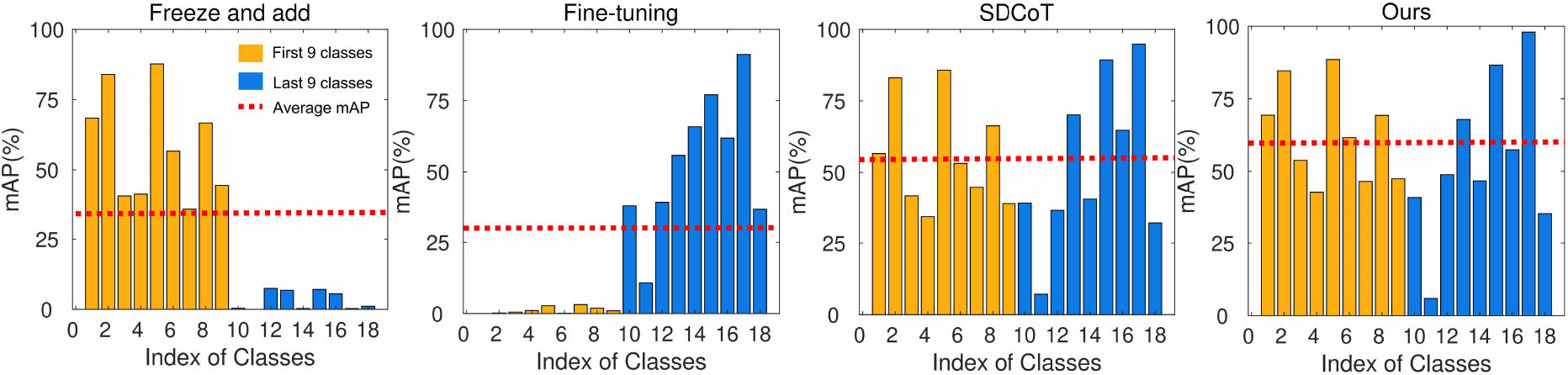}
	\vspace{-10pt} 
	\caption{Class-wise performance comparisons (mAP@0.25) on ScanNet dataset under the setting of 9 + 9.} 
	\label{fig: class-wise expreiments}
	\vspace{-20pt}
\end{figure*}

To handle the above problems, we propose the relation feature distillation to capture the responses relation between  $\mathbf{f}_i^t$ and $\mathbf{f}_i^{t-1}$.  We define our relation feature distillation loss function as:
\begin{align}
\label{eq: eq8}
{\mathcal{L}}_{RFD} = \sum_{\mathbf{x}_i^t\,\mathbf{x}_j^t \in \mathcal{D}^t }\frac{ \Vert Cos(\mathbf{f}_i^t,\mathbf{f}_j^t) - Cos(\mathbf{f}_i^{t-1},\mathbf{f}_j^{t-1})  \Vert_2}{C_{|\mathcal{D}^t|}^2},
\end{align}
where $C_{|\mathcal{D}^t|}^2$ means the total number of paired samples selected from training data $\mathcal{D}^t$.

At each training iteration, we compute a supervised loss $\mathcal{L}_S$ via following the loss function in VoteNet \cite{b6}. The supervised loss of our prompt guidance block ${\mathcal{L}}_{PGB}$ is also included in $\mathcal{L}_S$. Overall, the model $\Phi_i^{t}$ is optimized as follows:
  	  $$ 
{\mathcal{L}} = \alpha\mathcal{L}_S + \beta\mathcal{L}_{RDD} + \gamma\mathcal{L}_{RFD} +\xi\mathcal{L}_{Dis}, \eqno(9)$$  
where $\alpha, \beta,\gamma,\xi$ are hyperparameters in our experiments.

\section{EXPERIMENTS}

\subsection{Datasets and Evaluation metrics}

\textbf{Datasets.}
By following SDCoT \cite{b2}, we organize our experiments on SUN RGB-D \cite{b22} and ScanNet \cite{b27} datasets. We present the details for these two datasets as follows: 1) \textbf{SUN RGB-D} consists of 5,285 training samples for hundreds of object categories and has 5,050 samples to evaluate. In line with the standard evaluation protocol in SDCoT, we use the same 10 categories to report our performance. 2) \textbf{ScanNet} includes 1,201 samples for training and 312 samples for evaluating. Following VoteNet \cite{b6}, we generate our input point clouds by selecting vertices from the reconstructed meshes and gain the bounding boxes from the point-level labeling. We choose the same 18 categories to start our experiments.

\textbf{Evaluation metrics.} We use mean average precision (mAP@0.25) and recall in 3D object detection as our evaluation metrics. 
\subsection{Implementation Details}

For the hyper-parameters in our framework, we define the number of selected prompts $S$ as 10 and  $\zeta$ in Eq. 6 as 1.2. The weights of loss functions in Eq. 9  are designed as $\alpha=10$, $\beta=0.8$, $\gamma=1$, $\xi=1$. We adopt the same method in SDCoT to schedule the respective contributions of $\beta$, $\gamma$ and $\xi$. For our prompt guidance block, we randomly initialize prompts as learnable embeddings at the beginning of the first task and adopt $\mathcal{L}_{PGB}$ in Eq. 5 to optimize. After training on the  current task, we store these prompts in prompt pool, which are used to initialize prompts in the next task. The adam optimizer with the initial learning rate 0.001 is adopted to train all models. In addition, the adam optimizer is scheduled via the cosine annealing schedule.

Concretely, we design two baselines for class-incremental 3D object detection (\emph{i.e.} Freeze and add, Fine-tuning in Table. \textrm{I}). The method Freeze and add means that we freeze the model $\Phi^{t-1}$ to initialize the model $\Phi^{t}$. Then, a new classifier based on $\mathcal{C}^t$ is added and trained as the only learnable module. Fine-tuning denotes that  we fine-tune $\Phi^{t-1}$ (all parameters except the old classifier) with a new classifier on the training data $\mathcal{D}^t$.  In addition to the two baselines, we compare our I3DOD with SDCoT using the identical class order and task setting.

\begin{figure}[t]
	\centering
	\includegraphics[width=230pt, height=230pt]
	{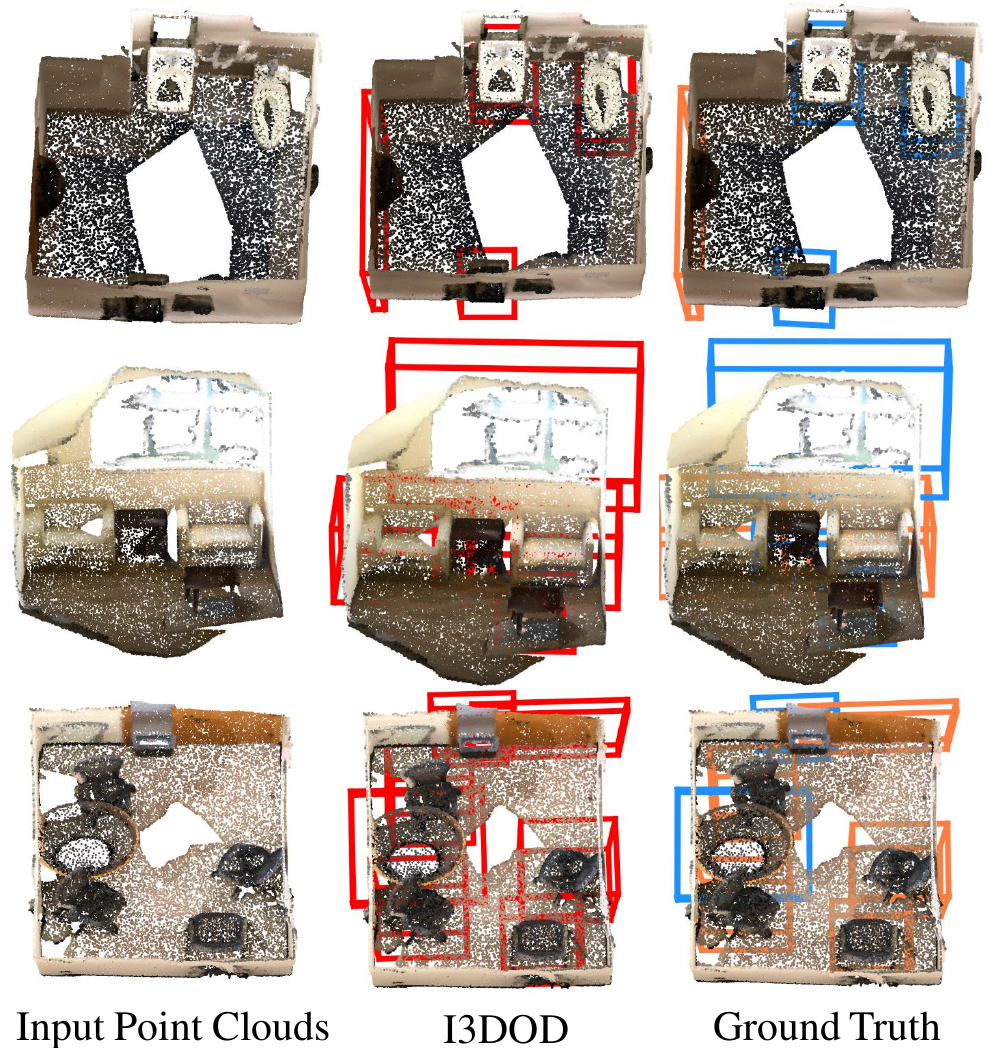}
	\vspace{-5pt}
	\caption{Qualitative results on ScanNet dataset. The left is the three original scenes. The middle is the scenes with bounding boxes generated by our I3DOD. The right is the scenes with ground truth bounding boxes. \textcolor{orange}{\textbf{Yellow}} and \textcolor{blue}{\textbf{blue}} denote the bounding boxes of old classes and new classes of ground truth,  and \textcolor{red}{\textbf{red}} represents the predict bounding boxes of our I3DOD.} 
	\label{fig: visual}
	\vspace{-17pt}
\end{figure}
\subsection{Quantitative Results}

    \textbf{SUN RGB-D:} As shown in Tabs.~\ref{tab: sunrgb}--\ref{tab: class-wise} and Fig.~\ref{fig: class-wise expreiments}, we evaluate the effectiveness of all experiments on SUN RGB-D. In this paper, \textbf{B} means the base task and \textbf{N} denotes the novel task (\emph{e.g}., 5+5 means that the base task consists of the first five classes and the other five classes belong to the novel task).Our model gains significant improvement over other methods by $0.9\%\sim22.0\%$ in terms of task average mAP@0.25. In the setting of 9+1, our I3DOD significantly exceeds the state-of-the-art method SDCoT by $2.9\%$, $8.5\%$ in terms of  mAP@0.25 and recall. 
   
    \textbf{ScanNet:} We report the performance of all comparison experiments in Tab.~\ref{tab: scannet} and visualize some evaluation results in Fig. 4. Compared with other methods, our I3DOD achieves solid improvement in all incremental settings. Furthermore, I3DOD outperforms SDCoT by $0.6\%$ in the base task and $4.1\%$ in the novel task.  All the experiments show that our I3DOD can tackle catastrophic forgetting from two aspects of category information comprehension and reliable representation distillation.

\subsection{Ablation Studies}

\begin{figure}[t]
	\centering
	\includegraphics[width=239pt, height=100pt]
	{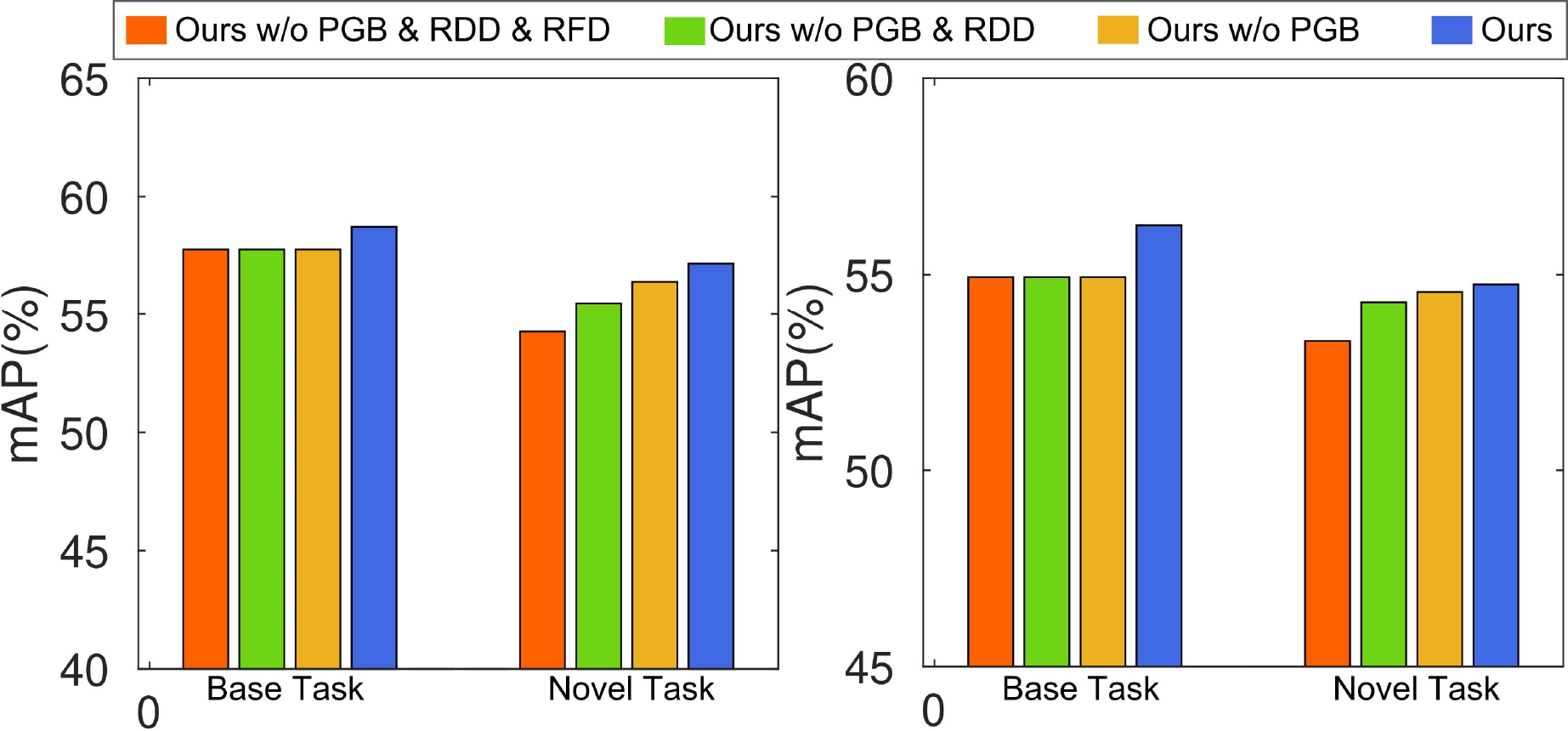}
	\caption{Ablation study results on SUN RGB-D dataset under the class setting of 5+5 (left) and 7+3 (right). } 
	\label{fig: ablation}
	\vspace{-15pt}
\end{figure}

As presented in Tab.~\ref{tab: sunrgb} and Fig.~\ref{fig: ablation}, we ablate the effectiveness of each module in our I3DOD. We compare our I3DOD with I3DOD eliminated our proposed modules one by one (\emph{i.e.,} PGB, RDD, RFD). Specifically, we notice that the performance of Ours w/o PGB degrades $0.8\%\sim2.3\%$ in terms of task average mAP@0.25, which show that our prompt guidance block can perform category information comprehension to gain identical improvement of new and old classes. Furthermore, Ours w/o PGB \& RDD is worse by $0.2\%\sim2.0\%$ in comparison with Ours w/o PGB, and achieves better performance compared with the baseline Ours w/o PGB \& RDD \& RFD, which reflects our distillation strategy achieves the significant effectiveness.

\section{CONCLUSIONS}

In this paper, we propose a novel I3DOD framework to tackle the challenge of catastrophic forgetting in class-incremental 3D object detection. In particular, the prompt guidance block is designed to capture the relationship
between the object localization information and high-level semantic information. We then point out that the responses of old detection model consist of positive responses and negative responses, and present our reliable dynamic distillation to filter out the negative knowledge of old model. Meanwhile, a relation feature distillation module is designed to the spatial positional relationship in feature space. Finally, we report our significant performance against baseline methods and verify the effectiveness of each module in I3DOD.

\addtolength{\textheight}{-12cm}   





\bibliographystyle{IEEEtran}
\bibliography{ref}{}

\end{document}